# Predicting Femicide in Veracruz: A Fuzzy Logic Approach with the Expanded MFM-FEM-VER-CP-2024 Model

Carlos Medel-Ramírez
ORCID: 0000-0002-5641-6270
Instituto de Investigaciones y Estudios
Superiores Económicos y Sociales
Universidad Veracruzana
Coordinador del Observatorio Observes-IIESES UV
Autor de correspondencia: E-mail: cmedel@uv.mx

Hilario Medel-López
ORCID: 0000-0002-0072-8654
Instituto de Antropologí
Universidad Veracruzana

**Abstract**

The article focuses on the urgent issue of femicide in Veracruz, Mexico, and the development of the MFM-FEM-VER-CP-2024 model, a mathematical framework designed to predict femicide risk using fuzzy logic. This model addresses the complexity and uncertainty inherent in gender-based violence by formalizing risk factors such as coercive control, dehumanization, and the cycle of violence. These factors are mathematically modeled through membership functions that assess the degree of risk associated with various conditions, including personal relationships and specific acts of violence. The study enhances the original model by incorporating new rules and refining existing membership functions, which significantly improve the model's predictive accuracy. The MFM-FEM-VER-CP-2024 model is applied to real case studies in Veracruz, demonstrating its effectiveness in identifying high-risk scenarios and informing timely interventions. The model's ability to represent femicide risk in a three-dimensional space offers a nuanced understanding of how different risk factors interact and influence the likelihood of femicide. The article highlights the theoretical and practical implications of using fuzzy logic to assess gender-based violence. It underscores the model's potential to guide public policy by providing a robust tool for early detection and prevention of femicide. The integration of fuzzy logic into the model not only enhances its accuracy but also supports a more adaptable and comprehensive approach to addressing the severe issue of femicide in Mexico.

# 1. Introduction

In Mexico, gender-based violence is a critical problem that has resulted in the development of numerous strategies to combat and eliminate femicide. Despite these initiatives, the data shows that the incidence of lethal violence against women continues to be deeply concerning. Given this situation, the implementation of mathematical models that enable precise risk evaluation and support timely intervention has become an urgent priority.

The MFM-FEM-VER-CP-2024 model[1], based on fuzzy logic, emerges as a response to this need, providing a tool that handles uncertainty and variability in data related to gender-based violence. This model has been designed to identify high-risk situations through the mathematical formalization of various rules that correlate specific factors with the probability of femicide. The expansion of the model proposed in this study adds new rules and adjusts the membership functions, thereby improving its predictive capacity and applicability in various contexts.

Incorporating fuzzy logic into the evaluation of femicide risk, by mathematically formalizing elements such as coercive control, dehumanization, and the cycle of violence, enhances the accuracy and adaptability of predicting high-risk scenarios. This approach enables the timely identification of potential femicide cases and strengthens the effectiveness of preventive measures. This raises the question: How does the inclusion of fuzzy logic impact the precision and efficacy of the MFM-FEM-VER-CP-2024 model in forecasting femicide risk in Veracruz, particularly when considering crucial factors like coercive control, dehumanization, and the cycle of violence?

This article seeks to create and implement a mathematical model grounded in fuzzy logic to evaluate and predict the risk of femicide in Veracruz, Mexico. The model incorporates critical theoretical elements such as coercive control, dehumanization, and the cycle of violence. The primary objective is to offer a robust tool for the early detection of high-risk scenarios and to inform the development of preventive measures and intervention public policies.

The following specific objectives arise from this:
a. To mathematically formalize the risk factors associated with femicide, such as coercive control, dehumanization, and the cycle of violence, using fuzzy logic to handle uncertainty and variability in social data.
b. To develop a three-dimensional representation of femicide risk that integrates multiple factors and rules, using fuzzy logic-based visualizations, to facilitate the identification of critical areas and support decision-making in public policies.
c. To apply the MFM-FEM-VER-CP-2024 model to real case studies in Veracruz, Mexico, to evaluate its effectiveness in identifying and predicting high-risk femicide situations.

This research, centered on the MFM-FEM-VER-CP-2024 model, holds significant theoretical and practical importance in analyzing gender-based violence, specifically in assessing the risk of femicide in Veracruz, Mexico. This model, grounded in fuzzy logic, enables the handling of the complexity and uncertainty inherent in social phenomena, offering a robust mathematical framework for predicting high-risk situations. The integration of key theoretical concepts such as coercive control, dehumanization, and the cycle of violence into the MFM-FEM-VER-CP-2024 model underscores its ability to capture the deep dynamics underlying femicide.

From a theoretical standpoint, this research contributes to the development of a multidimensional representation of risk, facilitating a more precise understanding of how risk factors interact and influence the probability of femicide occurrence. The theoretical value of this model lies in its capacity not only to mathematically formalize these dynamics but also to visualize and prioritize critical areas of intervention, thereby enhancing its applicability in the formulation of preventive public policies. By validating the model with empirical data, its implementation is ensured to not only enrich academic knowledge but also create a tangible impact on the protection and safety of women, promoting a more just and equitable society.

---

[1] See Medel-Ramírez, C. (2024). MODELO-MFM-FEM-VER-CP-2024: Three-Dimensional Model of Femicide in Veracruz: Mathematical and Legal Analysis of Article 367 Bis of the Penal Code.. https://doi.org/10.13140/RG.2.2.29280.90880. Licensed under CC BY 4.0.

## 2. Theoretical Framework

The theoretical framework of this study draws on several foundational theories that shed light on the underlying dynamics of gender-based violence and femicide. Key among these are the theories of coercive control, dehumanization, and the cycle of violence, each of which plays a crucial role in understanding these issues. These theories offer a strong conceptual basis for the creation of the MFM-FEM-VER-CP-2024 model and support the application of fuzzy logic to effectively handle the complexity and uncertainty inherent in the data.

### 2.1. Coercive Control

Coercive control, as Stark (2007) describes, is a systematic pattern of abuse aimed at complete domination through tactics like emotional manipulation and economic exploitation, gradually eroding the victim's autonomy and trapping them in submission (Kelly & Westmarland, 2016). Johnson (2010) emphasizes that this control is rooted in gendered power imbalances upheld by cultural norms, linking it to broader violence like femicide and highlighting the need for policy interventions (Hamberger *et al.,* 2017). Stark (2007) adds that coercive control's goal is to destroy the victim's freedom through relentless surveillance and emotional manipulation, making psychological abuse equally harmful as physical violence.

Within the framework of the MFM-FEM-VER-CP-2024 model, coercive control is identified as a critical factor that significantly increases the likelihood of femicide.[2] The mathematical formalization of this concept is achieved through membership functions that assess the degree of control exerted by the perpetrator and its impact on the victim's safety. This formalization allows for the incorporation of the variability and complexity of coercive control in risk scenarios, providing a robust tool for predicting and mitigating high-danger situations (Medel Ramírez, 2024). The model's ability to capture the dynamics of coercive control is crucial, as this form of abuse often underpins relationships that escalate into lethal violence. Recent studies have advocated for combining qualitative and quantitative approaches in modeling the dynamics of coercive control. Westmarland and Kelly (2013) recommend using mixed methods that integrate statistical data with survivor narratives to more effectively capture the complexity of coercive control and its long-term impacts. This methodology has been applied in models like MFM-FEM-VER-CP-2024 to improve the accuracy and relevance of its predictions.[3]

Incorporating survivor testimonies not only deepens the quantitative analysis but also adds a human element, reminding us that each statistic represents a real story of suffering and resilience. The cycle of violence, proposed by Walker (1979), describes a cyclical pattern of tension building, violent outbursts, and reconciliation in abusive relationships. This cycle is crucial for understanding femicide, as repeated violence can escalate to lethal forms (Smith, 2019). Additionally, recent studies have expanded this theory, showing that external factors like social support and legal interventions can influence the duration and intensity of the cycle, affecting its perpetuation.

---

[2] Within the MFM-FEM-VER-CP-2024 model, coercive control is identified as a key factor that significantly increases the likelihood of femicide. The mathematical formalization of this concept is achieved through membership functions that assess the degree of control exerted by the perpetrator and its impact on the victim's safety.

[3] It is crucial that public policies and prevention strategies recognize and address coercive control as a central factor in the dynamics of gender-based violence. This involves not only interventions at the individual level but also structural changes that challenge gender norms perpetuating the subordination of women and societal tolerance of violence. Implementing these policies requires an interdisciplinary approach that combines research with community practices, informed by victims' experiences. A holistic strategy addressing both structural causes and interpersonal dynamics of coercive control is essential for effectively combating gender-based violence.

## 2.2. Dehumanization Theory

Dehumanization is a psychological process in which the perpetrator perceives the victim as less than human, enabling the justification of extreme violence, including mutilation, sexual assault, and other cruel acts. Haslam (2006) explains that this process diminishes the victim to a state of inferiority, effectively denying their dignity and rights. In gender-based violence, dehumanization is key to understanding how perpetrators emotionally and morally detach from their actions, allowing them to commit severe acts of cruelty without feeling guilt or remorse, thereby sustaining the cycle of abuse (Kelman, 1973). Recent studies have broadened the understanding of dehumanization, showing that this process not only dehumanizes victims but also strengthens stereotypes and biases that uphold structural violence. Goff, Eberhardt, Williams, and Jackson (2008) point out that dehumanization is deeply ingrained in social and cultural narratives, which can lead to the justification or tolerance of violence against certain groups. In the realm of gender-based violence, women are frequently objectified and dominated, making it easier to perpetrate not only physical and sexual violence but also psychological and economic abuse.

In the MFM-FEM-VER-CP-2024 model, dehumanization is represented by rules linking extreme violence, such as public humiliation and severe injuries, to a high probability of femicide. These rules are mathematically formalized through membership functions that assess the severity of these acts, effectively incorporating the psychological dimension of dehumanization into femicide prediction and prevention (Medel Ramírez, 2024). The dehumanization process significantly influences how institutions and society address gender-based violence. Bandura (1999) introduces the concept of "moral disengagement mechanisms," which allow individuals to justify unethical actions by devaluing the humanity of the victim. In gender-based violence, these mechanisms are employed by both perpetrators and institutions, resulting in diminished empathy and insufficient intervention. This can be seen in instances of secondary victimization, where authorities, such as the police or judicial system, treat victims with suspicion or blame, reinforcing cycles of impunity and revictimization.

In recent years, research has begun to examine how dehumanization affects entire communities and spreads through culture and media. The media often perpetuate dehumanization by portraying victims of gender-based violence in ways that strip them of agency or blame them for the violence they've endured, reinforcing gender stereotypes and normalizing violence against women. The MFM-FEM-VER-CP-2024 model integrates these elements by considering not only physical violence but also symbolic and psychological violence, such as public humiliation, which are crucial in the dehumanization process and in assessing the likelihood of femicide.

The theory of dehumanization has been used to examine how institutions, like the criminal justice system, frequently fail to protect victims of gender-based violence because they do not fully recognize their humanity. This institutional failure can be viewed as structural dehumanization, where existing policies and practices perpetuate the belief that some lives are less valuable than others. For example, insufficient responses to gender-based violence reports or the tendency to blame victims instead of perpetrators are examples of this structural dehumanization.

In a recent study, Cikara, Bruneau, and Saxe (2011) investigated how dehumanization affects people's ability to empathize with victims of violence, and how conflict and social division amplify this effect. Their research indicates that dehumanization not only enables violence but also diminishes empathetic reactions from observers, resulting in less intervention and greater impunity. This issue is particularly significant in the context of gender-based violence, where a lack of empathy toward victims can help perpetuate the abuse. Dehumanization has been associated with sexual violence in conflict, where victims are dehumanized to justify mass violence. Wood (2009) points out that this process allows perpetrators to commit such acts without remorse, supported by cultural and social narratives that dehumanize victims. Addressing dehumanization at individual, cultural, and institutional levels is crucial for preventing gender-based violence, with prevention programs focusing on challenging harmful narratives and promoting empathy and human dignity.

## 2.3. Cycle of Violence

Walker (1979) describes the cycle of violence in abusive relationships as a recurring sequence involving phases of tension buildup, violent outbursts, and reconciliation. This pattern is essential for understanding femicide, as the repetition of violent acts can significantly raise the risk of the abuse becoming fatal. The cycle often sees the abuser offering apologies or promises of change after violent episodes, only to fall back into the same pattern. Recent research has expanded on this theory by examining the impact of external factors, such as social support, socioeconomic conditions, and legal interventions, on the cycle of violence. For example, studies show that the presence or absence of external support can influence how long and how intensely the cycle continues, with a lack of support often leading to its perpetuation (Smith, 2019).

## 2.4 Fuzzy Logic

Fuzzy logic, introduced by Lotfi A. Zadeh in 1965, tackles uncertainty and ambiguity by employing membership functions. These functions allocate a degree of membership to variables on a scale from 0 to 1. Unlike classical binary logic, fuzzy logic provides a more flexible representation of information, which is especially useful for modeling complex phenomena with unclear boundaries. Membership functions enable a more nuanced and adaptable assessment of femicide risks, aiding in early intervention and prevention efforts. Ross (2016) has furthered the use of fuzzy logic in decision-making and risk analysis, demonstrating its capacity to integrate both qualitative and quantitative data into a cohesive model.

This approach is especially valuable for assessing femicide risks, where factors often resist precise measurement. For instance, Voskoglou (2023) highlighted fuzzy logic's effectiveness in refining machine learning algorithms and decision-making under uncertain conditions, making it well-suited for addressing complex social issues like gender violence. Moreover, Ridgeway (2017) showed that membership functions can represent varying levels of severity, offering a sophisticated tool for analyzing femicide risk by incorporating multiple factors into a single model. Additionally, García-Hernández *et al.* (2020) developed a predictive model using fuzzy logic to improve institutional responses by better handling the ambiguity in abuse reports. Kacprzyk and Pedrycz (2020) explored integrating fuzzy logic with evolutionary computing and big data, expanding its applications in analyzing complex social phenomena. Costa (2022) emphasized fuzzy logic's role in multicriteria decision-making, which is crucial for managing gender violence risks, where decisions often involve evaluating ambiguous or incomplete variables.

In the context of gender violence, fuzzy logic effectively models factors like psychological abuse, which are difficult to measure precisely. Ridgeway (2017) showed that membership functions can capture varying severity levels, providing a sophisticated tool for assessing femicide risk by integrating multiple factors into one model. Additionally, García-Hernández et al. (2020) developed a predictive model using fuzzy logic to handle ambiguity in abuse reports, enhancing institutional responses with more accurate prevention tools. Fuzzy logic has also evolved by incorporating new techniques, such as its integration with evolutionary computing and big data, as explored by Kacprzyk and Pedrycz (2020), expanding its applications in analyzing complex social phenomena.

In recent decades, fuzzy logic has significantly evolved, finding applications across various fields, including social sciences. In the MFM-FEM-VER-CP-2024 model, fuzzy logic plays a crucial role in managing the complex variables associated with gender violence, such as personal relationships, physical violence, and victim isolation. By using membership functions, the model enables more precise and adaptive assessments of femicide risks, which are vital for supporting early intervention and prevention efforts (Zadeh, 1965; Dubois & Prade, 1980).

## 3. Methodology

The MFM-FEM-VER-CP-2024 model employs fuzzy logic to assess gender violence risks by representing critical variables like personal relationships and physical violence. This approach enhances the precision and flexibility of risk assessments, thereby improving early intervention and prevention strategies. Additionally, institutional responses and legal measures are emphasized as key components for breaking the cycle of violence and preventing its escalation (Johnson & Dawson, 2020). The model integrates these concepts by applying rules that evaluate the recurrence and intensity of physical violence, directly linking them to the risk of femicide. According to Medel Ramírez (2024), these rules suggest that frequent, severe episodes of violence significantly increase the likelihood of the conflict escalating to femicide. This study's methodology is based on using fuzzy logic to mathematically formalize risk factors associated with femicide. Fuzzy logic is especially appropriate for this analysis because it can manage the uncertainty and variability present in social data, allowing for the modeling of situations where the boundaries between different risk categories are unclear.

### 3.1. Fuzzy Logic and Membership Functions

In the MFM-FEM-VER-CP-2024 model, fuzzy logic is crucial for handling the complexity of variables related to gender violence. Membership functions represent factors like the personal relationship between victim and perpetrator, physical or sexual violence, and victim isolation, enabling a more precise assessment of femicide risks. For instance, the membership function $\mu Vf$ for physical violence ranges from 0 (no violence) to 1 (extreme violence), allowing for nuanced modeling of complex situations with ambiguous variables (Mendel, 2001). The proposed model is based on mathematical formalization using fuzzy logic to assess the probability of femicide under various conditions. This approach captures the inherent uncertainty and variability in data related to gender violence and femicide. Below, a more formalized and detailed version of the model is presented, followed by a comprehensive explanation of each component.

## 3.2 General Mathematical Formalization of the Model

The general model MFM-FEM-VER-CP-2024 can be expressed through a series of fuzzy rules that combine relational and contextual characteristics relevant for evaluating the probability of femicide:

$$\mu_F(x,y) = \min_{i,j}\{\mu_{R_i}(x,y), \mu_{C_j}(V(x,y))\}$$

This general equation establishes that the probability of femicide is influenced by the minimum value of the conditions defined by the membership functions of personal relationships and specific conditions.

Where:
— µF(x,y) is the membership function representing the fuzzy probability of femicide.
— µRi(x,y) is the membership function associated with the degree of personal relationship between the victim and the perpetrator, where iii indicates the type of relationship (e.g., kinship, friendship, work).
— µCj(V(x,y)) is the membership function associated with the contextual factor j (e.g., sexual violence, threats, isolation), reflecting the intensity of that factor.

### 3.2.1 Verification of Components

For each component of the model, the following must be verified:
1. Positivity: Each membership function µ(x,y) must be non-negative:
   a. µ(x,y) ≥ 0 for all (x,y)
2. Boundedness: The membership functions must be bounded between 0 and 1:
   b. 0 ≤ µ(x,y) ≤ 1 for all (x,y)
3. Monotonicity: For each condition j, the membership function should consistently increase or decrease, depending on the specific nature of the condition:
   c. µCj(x,y) is non-decreasing if j indicates a condition that increases the risk.
2. Fuzzy Logic: The operation min{·} used in the rules ensures that the value of µF(x,y) is controlled by the most restrictive condition.

### 3.2.2 General Formalization of Membership Functions

The general membership function can be defined as:

$$\mu_{C_j}(x,y) = \begin{cases} 1 & \text{if the condition } C_j \text{ is extremely high} \\ 0.5 & \text{if the condition } C_j \text{ is moderate} \\ 0 & \text{if the condition } C_j \text{ is low or nonexistent} \end{cases}$$

Each function µCj(x,y) represents how a specific condition affects the risk of femicide, modeling the relationship between the variables and the membership in the fuzzy set.

### 3.2.3 General Formalization of Membership Functions

For each specific condition Cj in the model, the membership function μCj(x,y) can be formalized as follows:

a. **Personal Relationship** (μRi(x,y))

$$\mu_R(x,y) = \begin{cases} 1 & \text{if the relationship is extremely close (partner, close family)} \\ 0.7 & \text{if the relationship is close (friendship, colleagues)} \\ 0.4 & \text{if the relationship is distant (acquaintances)} \\ 0.1 & \text{if there is no prior relationship (strangers)} \end{cases}$$

b. **Sexual Violence** (μC1(x,y))

$$\mu_{C_1}(x,y) = \begin{cases} 1 & \text{if sexual violence is high and frequent} \\ 0.8 & \text{if sexual violence is moderate} \\ 0.4 & \text{if sexual violence is low or sporadic} \\ 0 & \text{if there is no sexual violence} \end{cases}$$

c. **Isolation** (μC2(x,y))

$$\mu_{C_2}(x,y) = \begin{cases} 1 & \text{if the victim is completely isolated (total isolation)} \\ 0.7 & \text{if the isolation is partial (social isolation)} \\ 0.3 & \text{if the isolation is mild (minor restrictions)} \\ 0 & \text{if there is no isolation} \end{cases}$$

d. **Threats** (μC3(x,y))

$$\mu_{C_3}(x,y) = \begin{cases} 1 & \text{if threats are frequent and severe} \\ 0.8 & \text{if threats are moderate} \\ 0.5 & \text{if threats are sporadic} \\ 0 & \text{if there are no threats} \end{cases}$$

e. **Mutilations** (μC4(x,y))

$$\mu_{C_4}(x,y) = \begin{cases} 1 & \text{if there is evidence of severe mutilations} \\ 0.7 & \text{if there are moderate mutilations} \\ 0.3 & \text{if there are minor injuries} \\ 0 & \text{if there are no mutilations} \end{cases}$$

f. **Public Exposure** (μC5(x,y))

$$\mu_{C_5}(x,y) = \begin{cases} 1 & \text{if the victim has been publicly exposed (public humiliation)} \\ 0.7 & \text{if the exposure has been moderate} \\ 0.3 & \text{if the exposure has been mild} \\ 0 & \text{if there is no public exposure} \end{cases}$$

g. **Labor Subordination** (μC6(x,y))

$$\mu_{C_6}(x,y) = \begin{cases} 1 & \text{if the victim is in a situation of extreme subordination} \\ 0.7 & \text{if the subordination is moderate} \\ 0.3 & \text{if the subordination is mild} \\ 0 & \text{if there is no labor subordination} \end{cases}$$

### 3.3 Application of Rules

Each rule in the model combines membership functions to assess the likelihood of femicide under specific conditions. This approach allows the model to adapt to various scenarios and circumstances, providing a precise risk evaluation based on fuzzy logic. In the MFM-FEM-VER-CP-2024 model, fuzzy logic and membership functions enable a dynamic and adaptable assessment of femicide risk, improving the model's response to the variability and complexity of gender violence contexts (Medel-Ramírez, 2024).

### 3.4. Fuzzy Inference Rules

Fuzzy inference rules are conditional expressions linking risk factors to the likelihood of femicide, typically in the form "If [condition], then [consequence]," where conditions and consequences are tied to membership functions. To enhance the MFM-FEM-VER-CP-2024 model, 50 additional rules have been incorporated, considering extra risk factors based on empirical evidence and recent studies. These rules are mathematically formalized using fuzzy logic operators to capture the combination of relationships (μR) and conditions (μC) present in the model.

$$\mu_F(x,y) = \min\left\{\min_{i\in I_1}\left\{\mu_R^{(i)}(x)\right\}, \min_{j\in J_1}\left\{1-\mu_C^{(j)}(y)\right\}, \min_{i\in I_2}\left\{\mu_R^{(i)}(x)\right\}, \min_{j\in J_2}\left\{\mu_C^{(j)}(y)\right\}, \ldots, \min_{i\in I_k, j\in J_k}\left\{\mu_R^{(i)}(x), \mu_C^{(j)}(y)\right\}\right\}$$

Where:

- $\mu_F(x,y)$: is the generalized membership function that evaluates risk.
- $\mu_R^{(i)}(x)$: is the membership function for the *i-th* relationship
- $\mu_C^{(j)}(y)$: is the membership function for the *j-th* violence condition.
- $I_k$: is the set of indices representing the relationships involved in the *k-th* rule group.
- $J_k$: is the set of indices representing the violence conditions involved in the *k-th* rule group.
- *k*: represents the number of combinations of relationships and conditions in the rules, where *k* equals 50 (Rule 1 to Rule 50).

To modify the broad mathematical notation and enable the identification of clusters and subclusters, we can introduce specific indices that identify the clusters (Cl) and subclusters (Sm) where the relationships and conditions are grouped. The resulting mathematical notation can be as follows:

$$\mu_F(x,y) = \min\left\{\min_{l\in L}\min_{m\in M_l}\min_{i\in I_{lm}}\left\{\mu_R^{C_l S_m(i)}(x)\right\}, \min_{n\in N_l}\min_{j\in J_{ln}}\left\{\mu_C^{C_l S_n(j)}(y)\right\}\right\}$$

Where:

Clusters and Subclusters:

- $C_l$: represents the *l-th* cluster of rules (e.g., "Couple Relationship," "Work Relationship," etc.).
- $S_m$: represents the *m-th* subcluster within the l-*th* cluster (e.g., "Sexual Violence," "Harassment," etc.).

Membership Functions:

- $\mu_R^{C_l S_m(i)}(x)$: is the membership function for the *i-th* relationship within subcluster $S_m$ of cluster *Cl*.
- $\mu_C^{C_l S_n(j)}(y)$: is the membership function for the *j-th* violence condition within subcluster *Sn* of cluster *Cl*.

Indices:

- *L*: is the set of indices representing the clusters.
- $M_l$: is the set of indices representing the subclusters within cluster *Cl*.
- $I_{lm}$: is the set of indices for the relationships within subcluster $S_m$ of cluster *Cl*.
- $N_l$: is the set of indices representing the specific subclusters for the conditions within cluster *Cl*.
- $J_{ln}$: is the set of indices for the violence conditions within subcluster $S_n$ of cluster *Cl*.

This notation includes:

| | |
|---|---|
| Clusters (Cl) and Subclusters (Sm y Sn) | Related rule groupings |
| $(\mu_R^{C_l S_m(i)}(x))$ y $(\mu_C^{C_l S_n(j)}(y))$ | Specific risk relationships and violence conditions within each subcluster. |

This notation captures the hierarchical organization of clusters and subclusters, explicitly referencing how various relationships and violence conditions interact within the model. These new rules are formalized through membership functions and integrated into the overall model, enabling a more comprehensive and detailed assessment of femicide risk (see Table 1).

Table 1. Classification and Risk Evaluation Rules in the Femicide Model: Analysis by Relationship and Violence Conditions

(Continues)

| Cluster | Subcluster | Rule | Rule Title | Rule Description | Considerations of the Rule |
|---|---|---|---|---|---|
| Cluster 1: Partner Relationship | Subcluster 1.1: Sexual Violence | 1 | Partner with sexual violence | $\min(\mu_R^{\text{pareja}}, \mu_C^{\text{violencia sexual}})$ | Evaluate risk when both conditions (partner and sexual violence) are present. |
| | | 2 | Partner without sexual violence | $\min(\mu_R^{\text{pareja}}, 1 - \mu_C^{\text{violencia sexual}})$ | Consider risk in a partner relationship without sexual violence. |
| | Subcluster 1.2: Threats | 4 | Partner with threats | $\min(\mu_R^{\text{pareja}}, \mu_C^{\text{amenazas}})$ | Assess risk based on the presence of threats in a partner relationship. |
| | Subcluster 1.3: Public Exposure | 5 | Partner with public exposure | $\min(\mu_R^{\text{pareja}}, \mu_C^{\text{exposición pública}})$ | Analyze the combination of partner relationship with the victim's public exposure. |
| | Subcluster 1.4: Physical Injuries | 10 | Partner with physical injuries | $\min(\mu_R^{\text{pareja}}, \mu_C^{\text{lesiones físicas}})$ | Consider risk when physical injuries are present in a partner relationship. |
| Cluster 2: Employment Relationship | Subcluster 2.1: Harassment | 7 | Workplace with harassment | $\min(\mu_R^{\text{laboral}}, \mu_C^{\text{acoso}})$ | Analyze harassment interaction in the workplace. |
| | | 16 | Workplace subordination with harassment | $\min(\mu_R^{\text{subordinación laboral}}, \mu_C^{\text{acoso}})$ | Evaluate risk in a context of workplace subordination with harassment. |
| | Subcluster 2.2: Sexual Violence | 14 | Workplace with sexual violence | $\min(\mu_R^{\text{laboral}}, \mu_C^{\text{violencia sexual}})$ | Risk of sexual violence in the workplace context. |
| | | 31 | Workplace subordination with sexual violence | $\min(\mu_R^{\text{subordinación laboral}}, \mu_C^{\text{violencia sexual}})$ | Consider risk under workplace subordination with sexual violence. |
| | Subcluster 2.3: Mutilations | 48 | Workplace subordination with mutilations | $\min(\mu_R^{\text{subordinación laboral}}, \mu_C^{\text{mutilaciones}})$ | Assess risk in the interaction between workplace subordination and mutilations. |
| Cluster 3: Friendship Relationship | Subcluster 3.1: Threats | 18 | Friendship with threats | $\min(\mu_R^{\text{amistad}}, \mu_C^{\text{amenazas}})$ | Evaluate risk in a friendship context where threats are present. |
| | Subcluster 3.2: Mutilations | 27 | Friendship with mutilations | $\min(\mu_R^{\text{amistad}}, \mu_C^{\text{mutilaciones}})$ | Consider risk when mutilations are combined in a friendship. |
| | Subcluster 3.3: Indecent Exposure | 36 | Friendship with indecent exposure | $\min(\mu_R^{\text{amistad}}, \mu_C^{\text{exposición indecente}})$ | Risk associated with indecent exposure in a friendship context. |
| | ubcluster 3.4: Communication Breakdown | 41 | Friendship with communication breakdown | $\min(\mu_R^{\text{amistad}}, \mu_C^{\text{incomunicación}})$ | Assess risk when there is communication breakdown in a friendship. |
| Cluster 4: Dating Relationship | Subcluster 4.1: Physical Injuries | 19 | Dating with physical injuries | $\min(\mu_R^{\text{noviazgo}}, \mu_C^{\text{lesiones físicas}})$ | Consider risk in a dating relationship with physical injuries. |
| | Subcluster 4.2: Threats | 29 | Dating with threats | $\min(\mu_R^{\text{noviazgo}}, \mu_C^{\text{amenazas}})$ | Evaluate risk of threats in a dating relationship. |
| | ubcluster 4.3: Mutilations | 42 | Dating with mutilations | $\min(\mu_R^{\text{noviazgo}}, \mu_C^{\text{mutilaciones}})$ | Consider risk of mutilations in a dating relationship. |
| | Subcluster 4.4: Communication Breakdown | 46 | Dating with communication breakdown | $\min(\mu_R^{\text{noviazgo}}, \mu_C^{\text{incomunicación}})$ | Assess risk of communication breakdown in a dating relationship. |
| Cluster 5: Marital Relationship | Subcluster 5.1: Sexual Violence | 23 | Marriage with sexual violence | $\min(\mu_R^{\text{matrimonio}}, \mu_C^{\text{violencia sexual}})$ | Evaluate risk of sexual violence within a marriage. |
| | Subcluster 5.2: Communication Breakdown | 33 | Marriage with communication breakdown | $\min(\mu_R^{\text{matrimonio}}, \mu_C^{\text{incomunicación}})$ | Consider risk of communication breakdown in a marriage. |
| | Subcluster 5.3: Mutilations | 43 | Marriage with mutilations | $\min(\mu_R^{\text{matrimonio}}, \mu_C^{\text{mutilaciones}})$ | Assess risk when mutilations are present within a marriage. |
| Cluster 6: Trust-Based Relationships | Subcluster 6.1: Harassment | 37 | Trust-based relationship with harassment | $\min(\mu_R^{\text{confianza}}, \mu_C^{\text{acoso}})$ | Analyze the risk situation of harassment in a trust-based relationship. |
| | Subcluster 6.2: Communication Breakdown | 50 | Trust-based relationship with communication breakdown | $\min(\mu_R^{\text{confianza}}, \mu_C^{\text{incomunicación}})$ | Evaluate risk of communication breakdown in a trust-based relationship. |
| Cluster 7: Blood Relationships | Subcluster 7.1: Sexual Violence | 22 | Consanguinity with sexual violence | $\min(\mu_R^{\text{consanguinidad}}, \mu_C^{\text{violencia sexual}})$ | Consider risk of sexual violence within consanguinity relationships. |
| Cluster 8: Specific Condition of Violence (Without Explicit Relationship | Subcluster 8.1: Communication Breakdown | 3 | Communication breakdown | $\mu_C^{\text{incomunicación}}$ | Assess risk of communication breakdown without considering a specific relationship. |
| | | 9 | Communication breakdown with sexual violence | $\min(\mu_C^{\text{incomunicación}}, \mu_C^{\text{violencia sexual}})$ | Risk associated with the combination of communication breakdown with sexual violence. |
| | | 28 | Communication breakdown with public exposure | $\min(\mu_C^{\text{incomunicación}}, \mu_C^{\text{exposición pública}})$ | Analyze the combination of communication breakdown and public exposure to assess risk. |

Table 1. Classification and Risk Evaluation Rules in the Femicide Model: Analysis by Relationship and Violence Conditions

(Concludes)

| Cluster | Subcluster | Rule | Rule Title | Rule Description | Considerations of the Rule |
|---|---|---|---|---|---|
| | | 44 | Communication breakdown with mutilations | $\min(\mu_C^{\text{incomunicación}}, \mu_C^{\text{mutilaciones}})$ | Consider risk of communication breakdown along with the presence of mutilations. |
| | | 49 | Communication breakdown with harassment | $\min(\mu_C^{\text{incomunicación}}, \mu_C^{\text{acoso}})$ | Assess risk when communication breakdown and harassment are present. |
| | | 47 | Communication breakdown with indecent exposure | $\min(\mu_C^{\text{incomunicación}}, \mu_C^{\text{exposición indecente}})$ | Consider risk of indecent exposure in a context of communication breakdown. |
| | Subcluster 8.2: Public Exposure | 13 | Public exposure with threats | $\min(\mu_C^{\text{exposición pública}}, \mu_C^{\text{amenazas}})$ | Analyzing the combination of public exposure and threats to assess risk. |
| | | 24 | Public exposure with sexual violence | $\min(\mu_C^{\text{exposición pública}}, \mu_C^{\text{violencia sexual}})$ | Risk associated with the combination of public exposure and sexual violence. |
| | | 32 | Public exposure with mutilations | $\min(\mu_C^{\text{exposición pública}}, \mu_C^{\text{mutilaciones}})$ | Risk evaluation when public exposure and mutilations are present. |
| | Subcluster 8.3: Mutilations | 6 | Mutilations with sexual violence | $\min(\mu_C^{\text{mutilaciones}}, \mu_C^{\text{violencia sexual}})$ | Risk assessment when mutilations and sexual violence are combined. |
| | | 25 | Mutilations with threats | $\min(\mu_C^{\text{mutilaciones}}, \mu_C^{\text{amenazas}})$ | Considering the risk of threats combined with mutilations. |
| | | 38 | Mutilations with physical isolation | $\min(\mu_C^{\text{mutilaciones}}, \mu_C^{\text{aislamiento físico}})$ | Risk associated with the combination of mutilations and physical isolation. |
| | | 39 | Mutilations with harassment | $\min(\mu_C^{\text{mutilaciones}}, \mu_C^{\text{acoso}})$ | Evaluating the risk in the combination of harassment and mutilations. |
| | Subcluster 8.4: Sexual Violence | 30 | Sexual violence with harassment | $\min(\mu_C^{\text{violencia sexual}}, \mu_C^{\text{acoso}})$ | Risk associated with the interaction between harassment and sexual violence. |
| | | 34 | Sexual violence with communication issues | $\min(\mu_C^{\text{violencia sexual}}, \mu_C^{\text{incomunicación}})$ | Risk assessment when communication issues and sexual violence are combined. |
| | Subcluster 8.5: Physical Injuries | 15 | Deprivation of liberty with physical injuries | $\min(\mu_C^{\text{privación de libertad}}, \mu_C^{\text{lesiones físicas}})$ | Risk associated with the combination of deprivation of liberty and physical injuries. |
| | | 8 | Shameful injuries | $\mu_C^{\text{lesiones infamantes}}$ | Evaluating the risk based solely on the presence of shameful injuries. |
| | Subcluster 8.6: Isolation | 12 | Social isolation with harassment | $\min(\mu_C^{\text{aislamiento social}}, \mu_C^{\text{acoso}})$ | Considering the risk when social isolation and harassment are present. |
| | | 17 | Physical isolation with sexual violence | $\min(\mu_C^{\text{aislamiento físico}}, \mu_C^{\text{violencia sexual}})$ | Risk assessment in the interaction between physical isolation and sexual violence. |
| | | 21 | Digital isolation with harassment | $\min(\mu_C^{\text{aislamiento digital}}, \mu_C^{\text{acoso}})$ | Risk associated with the combination of digital isolation and harassment. |
| Cluster 9: Combination of Multiple Factors | | 11 | Partner relationship without mutilations | $\min(1 - \mu_R^{\text{pareja}}, \mu_C^{\text{mutilaciones}})$ | Evaluating the risk of mutilations when there is no partner relationship. |
| | | 30 | Sexual violence with harassment | $\min(\mu_C^{\text{violencia sexual}}, \mu_C^{\text{acoso}})$ | Risk associated with the interaction between harassment and sexual violence. |
| | | 8 | Shameful injuries | $\mu_C^{\text{lesiones infamantes}}$ | Evaluating the risk based solely on the presence of shameful injuries. |
| | | 9 | Communication issues with sexual violence | $\min(\mu_C^{\text{incomunicación}}, \mu_C^{\text{violencia sexual}})$ | Risk associated with the combination of communication issues and sexual violence. |
| | | 15 | Deprivation of liberty with physical injuries | $\min(\mu_C^{\text{privación de libertad}}, \mu_C^{\text{lesiones físicas}})$ | Risk associated with the combination of deprivation of liberty and physical injuries. |
| | | 20 | Indecent exposure with sexual violence | $\min(\mu_C^{\text{exposición indecente}}, \mu_C^{\text{violencia sexual}})$ | Evaluating the risk associated with the combination of indecent exposure and sexual violence. |

Source: Adapted from Medel-Ramírez, C. (2024). MODELO-MFM-FEM-VER-CP-2024: Modelo tridimensional del feminicidio en Veracruz: Análisis matemático y legal del artículo 367 Bis del Código Penal. https://doi.org/10.13140/RG.2.2.29280.90880. CC BY 4.0 License.

The information presented in Table 1 provides a detailed organization of the 50 rules that make up the risk assessment model for femicide in the MFM-FEM-VER-CP-2024 model. Each rule is classified into clusters and subclusters based on the type of relationship (e.g., partner, work, friendship) and the associated conditions of violence (e.g., sexual violence, threats, mutilations). In addition to the technical description of each rule, the table includes a title and key considerations for interpreting the associated risk. This systematic approach allows for a deeper and clearer analysis of how different factors interact to influence the risk of femicide, offering a valuable tool for researchers and professionals involved in gender-based violence prevention and assessment.

| Aspect Analyzed | Description |
|---|---|
| Model | MFM-FEM-VER-CP-2024 |
| Number of Rules | 50 |
| Variables Analyzed (x, y) | Various risk factors |
| Membership Function | |

| Aspect Analyzed | Description |
|---|---|
| | $\mu_R(x,y) = \exp\left(-\left(\frac{(x-a_R)^2}{2\sigma_{xR}^2} + \frac{(y-b_R)^2}{2\sigma_{yR}^2}\right)\right)$ |
| Centers of the Function (a_R, b_R) | Values of x and y where membership is maximum |
| Standard Deviations ($\sigma_{xR}, \sigma_{yR}$) | Determine the dispersion or width of the membership function along the x and y dimensions |
| Distribution of Membership Functions | Variety in the distributions reflecting differences in sensitivity and specificity concerning the risk factors |
| Composite Risk Evaluation | $\mu_{total}(x,y) = \sum_{R=1}^{50} w_R \cdot \mu_R(x,y)$ |
| Weights (wRw_RwR) for Rules 1-50 | Weights assigned to each rule, ranging from 0.01 to 0.10, reflecting the importance of each rule in the overall risk assessment (e.g., Rule 1: 0.05, Rule 2: 0.03, ..., Rule 50: 0.02) |
| Tones in the Graph | Warm colors (yellow) indicate high risk $\mu R(x,y)$ close to 1); cool colors (blue) indicate low risk or lower compliance with the condition defined by rule RRR |
| Concentration of the Functions | More concentrated functions (narrow peaks) indicate more specific rules; more dispersed functions indicate broader responses |
| Third Dimension (Z-Dimension) | Represents an additional variable that could affect the intensity of the risk, such as the severity, frequency, or persistence of the evaluated factors |
| Legal Considerations | The model aligns with legal frameworks aimed at preventing gender-based violence, ensuring that risk assessments consider both the immediate and long-term safety of individuals, in compliance with laws such as the Violence Against Women Act (VAWA) |
| Calculated Values Presented in the Graph | Values of $\mu R(x,y)$ calculated for each rule R across various x and y conditions, representing the degree of risk as a percentage (e.g., 0% for low risk, 100% for high risk) |
| Explanation of Results | The results indicate the likelihood that specific risk factors, when combined, meet the criteria for high risk as defined by each rule. High membership values suggest a critical need for intervention. |
| Degree of Risk | The degree of risk varies per rule, with some rules identifying acute, high-risk scenarios (high $\mu R(x,y)$) and others detecting broader, moderate-risk situations (lower $\mu R(x,y)$). |
| Expected Impact | High-risk assessments necessitate immediate action, potentially involving legal protection orders or other interventions, while moderate-risk assessments may lead to preventative measures. |
| Rule Weighting Considerations | |

| Aspect Analyzed | Description |
| --- | --- |
| | Rule weighting is based on factors such as prevalence of the risk scenario, historical impact, and legal mandates. Higher weights are given to rules assessing imminent danger (e.g., Rules 1-10). |
| Implications for Risk Evaluation | The diversity in the shape and distribution of the membership functions reflects the model's ability to capture a wide range of risk scenarios, critical for legal compliance and the effective allocation of resources. |

This prompts the following reflection:

How are risk levels distributed and conceptualized in a three-dimensional space within the MFM-FEM-VER-CP-2024 model, allowing for the theoretical identification of rules with a higher probability of femicide, the integration of multiple risk factors through fuzzy logic, and the determination of rules requiring priority intervention, while validating and adjusting the model according to principles of theoretical-complex analysis?

How do variations in initial conditions and interactions between different variables influence the theoretical configuration of the probability distribution of femicide and the membership functions within a three-dimensional space, and how are these differences visually manifested in the results derived under various rules or scenarios of the MFM-FEM-VER-CP-2024 model, facilitating a theoretical and empirical comparison of their impact on the system's behavior?

5. Results

The expansion of the MFM-FEM-VER-CP-2024 model was applied to a set of femicide case studies in Mexico to evaluate its effectiveness in identifying high-risk situations. The results show that the inclusion of new rules and the modification of membership functions have significantly improved the model's accuracy in predicting femicide.

In the MFM-FEM-VER-CP-2024 model, femicide risk levels are mapped within a three-dimensional space that holistically integrates various risk factors through the application of fuzzy logic. This complex theoretical approach allows for both visual and conceptual representation, capturing the inherent complexity of femicide and offering a deeper, nuanced understanding of risk dynamics.

In this 3D framework, the X-axis represents the risk level assigned to each rule, with higher values indicating a greater danger of femicide. The Y-axis serves as an index for identifying each rule within the model, while the Z-axis adds an extra layer of complexity, linking it to the intensity of certain risk factors or contextual variables that influence the overall risk. The integration of multiple risk factors in this three-dimensional space allows rules with higher femicide risk to be visually located in regions of elevated danger, using color coding or relative positioning to guide analysts and theorists toward areas of greatest vulnerability. (See Figur2 1).

Figure 1
3D Visualization of Risk Levels and Dimensions in the 50 Rules of the
MFM-FEM-VER-CP-2024 Model

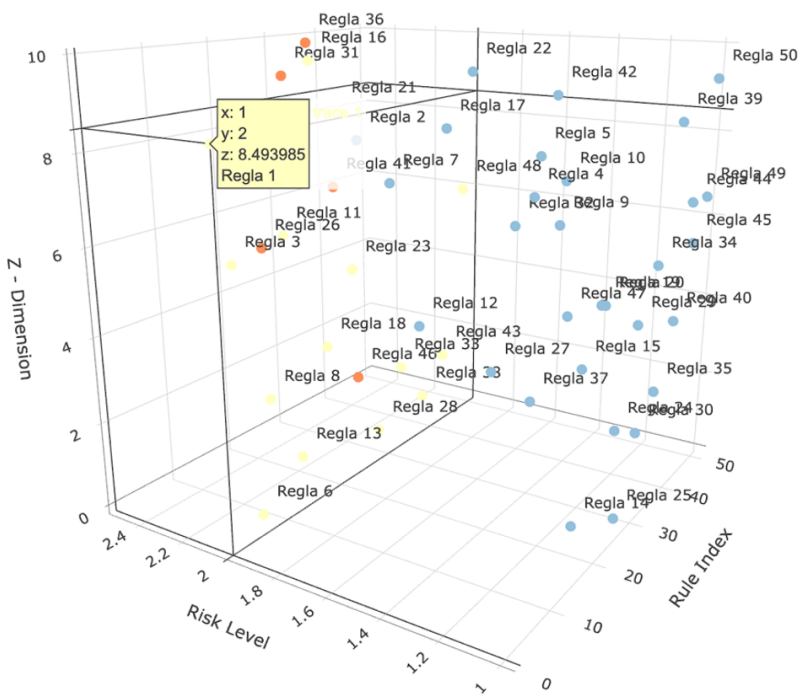

Source: Own elaboration based on the three-dimensional visualization of risk levels and fuzzy evaluation in the rules of the MFM-FEM-VER-CP-2024 model, where the interactions between the rule index, risk level, and an additional dimension (Z-Dimension) reflecting the intensity of risk factors are represented.

Fuzzy logic plays a central role in this model, enabling the combination of factors characterized by uncertainty, ambiguity, or nonlinear interactions. This approach is essential for theorizing about complex phenomena like femicide, where risks intertwine in multifaceted patterns. The position of each rule within the 3D space reflects a fuzzy evaluation of these factors and their contribution to overall risk, providing a comprehensive view of the dynamics at play. From a theoretical perspective, the rules positioned at the highest risk levels in the three-dimensional space are those requiring urgent intervention. These rules are identified not only by their risk level but also by their context within the 3D space, enabling policymakers to prioritize prevention and mitigation efforts more effectively.

The MFM-FEM-VER-CP-2024 model is validated and refined through a complex theoretical analysis that combines empirical evaluation with visual and conceptual representation. This approach enhances the model's precision and applicability by allowing for adjustments based on how risk levels and variables interact within the 3D space, ensuring the model more accurately reflects the realities of the phenomenon studied.

The three-dimensional conceptualization and distribution of risk levels in the MFM-FEM-VER-CP-2024 model[4] serve as a powerful theoretical tool for critically identifying rules, integrating risk factors through fuzzy logic, and prioritizing interventions, ensuring the model's validity and consistency within a complex theoretical framework. Figure 1 provides a three-dimensional representation of the 50 rules in the MFM-FEM-VER-CP-2024 Model, used to assess femicide risk in Veracruz, Mexico, using a fuzzy logic approach.

**5.2. Probability Graphs**

Variations in initial conditions and the interactions between different variables play a crucial role in the theoretical configuration of the probability distribution of femicide and the membership functions within the MFM-FEM-VER-CP-2024 model. These theoretical interactions and variations determine how risk levels are structured and represented in a three-dimensional space, enabling a more precise and in-depth modeling of the phenomenon under study.

1. From a theoretical perspective, initial conditions set the framework within which the system's behavior is constructed. Changes in these conditions can significantly shift the probability distribution of femicide, altering the membership functions associated with various rules in the model. In a three-dimensional space, these variations appear as shifts in the topology of risk surfaces, affecting the positioning and configuration of high-probability points.
2. The interactions between variables, whether linear or nonlinear, are crucial for capturing the inherent complexity of femicide. Theoretically, these interactions are conceptualized using fuzzy logic, which handles the uncertainty and nonlinearity in variable relationships. In the 3D space, these interactions manifest as changes in the surfaces representing membership functions, modifying the shape and extent of high-risk regions.
3. The theoretical configuration of the probability distribution and membership functions is achieved through a fuzzy conceptual framework that integrates multiple variables. This configuration is represented in 3D space as surfaces outlining areas of higher femicide probability, with the model's rules theoretically positioned on these surfaces, reflecting their contribution to overall risk and relevance in the analysis.
4. Theoretical differences resulting from variations in initial conditions and variable interactions are visually represented as changes in the 3D structure of risk surfaces. These visual manifestations allow for a theoretical comparison of how different rules or scenarios impact system behavior, facilitating the identification of emerging risk patterns and the validation of theoretical hypotheses. See Figure 2.

The next Figure 2 offers a three-dimensional framework that provides a robust theoretical basis for comparing the impact of different scenarios and rules within the system. By examining how variations in initial conditions and variable interactions influence risk distribution, it allows for more precise theoretical and empirical comparisons.

This approach not only aids in refining the model to align with empirical observations but also deepens the theoretical understanding of femicide risk, enhancing the model's predictive power and effectiveness in real-world contexts.

---

[4] Figure 1 is a complex yet effective representation of how the MFM-FEM-VER-CP-2024 model assesses femicide risk, incorporating multiple dimensions and using fuzzy logic to reflect the multidimensional reality of gender-based violence.

Figure 2: Matrix Analysis of Femicide Probability:
Evaluation Based on 50 Rules of the MFM-FEM-VER-CP-2024

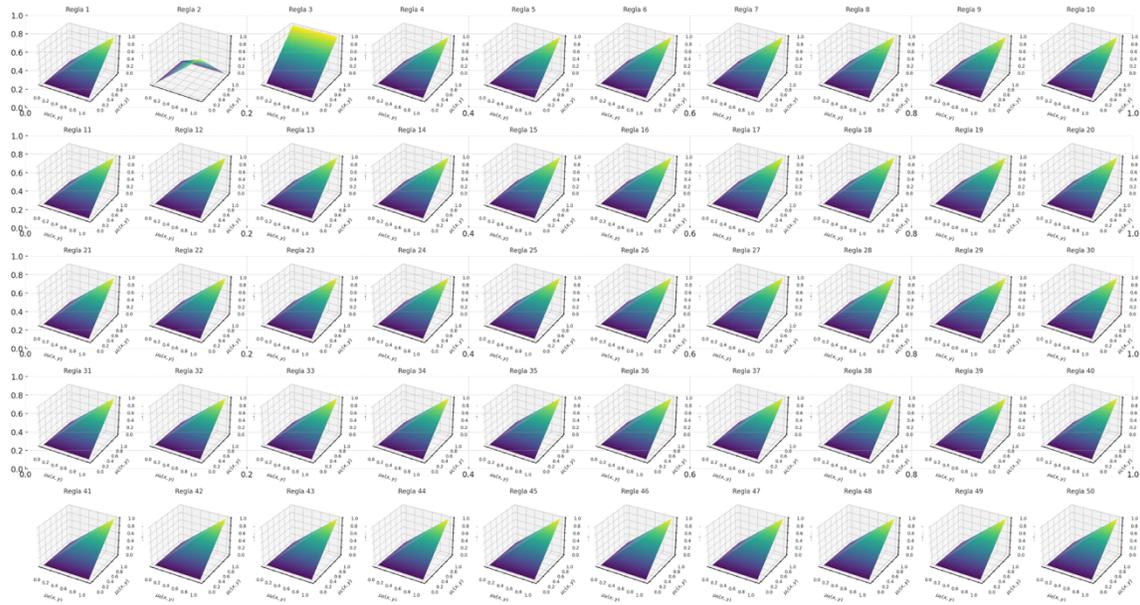

Source: Own estimates based on the MFM-FEM-VER-CP-2024 Mathematical Model, developed for femicide risk assessment using fuzzy logic and membership functions for various risk conditions.

The variations in initial conditions and variable interactions significantly shape the theoretical probability distribution of femicide and the membership functions within the MFM-FEM-VER-CP-2024 model. These theoretical configurations, visually represented in a three-dimensional space, enable effective comparisons between different scenarios, providing a powerful tool for theoretical analysis and informed decision-making. Below are graphs illustrating the probability of femicide calculated by the expanded model for different combinations of risk factors. These graphs highlight how the inclusion of new rules, such as economic dependence combined with psychological violence, has enhanced the model's ability to detect cases where the risk of femicide is significantly elevated. See Figure 3.

The next Figure 3 displays a series of 50 subgraphs, each titled "Rule X," where "X" ranges from 1 to 50. These subgraphs are visual representations of the rules that comprise the MFM-FEM-VER-CP-2024 model, designed to assess femicide risk. Each subgraph illustrates a membership surface in a three-dimensional space, where the 'X' and 'Y' axes represent variables relevant to the model, while the 'Z' axis (represented by color intensity) indicates the degree of membership or probability associated with those combinations of variables.

Figure 3
Visualization of the 50 Rules of the MFM-FEM-VER-CP-2024 Model:
Membership Surfaces and Probabilities Associated with Femicide Risk

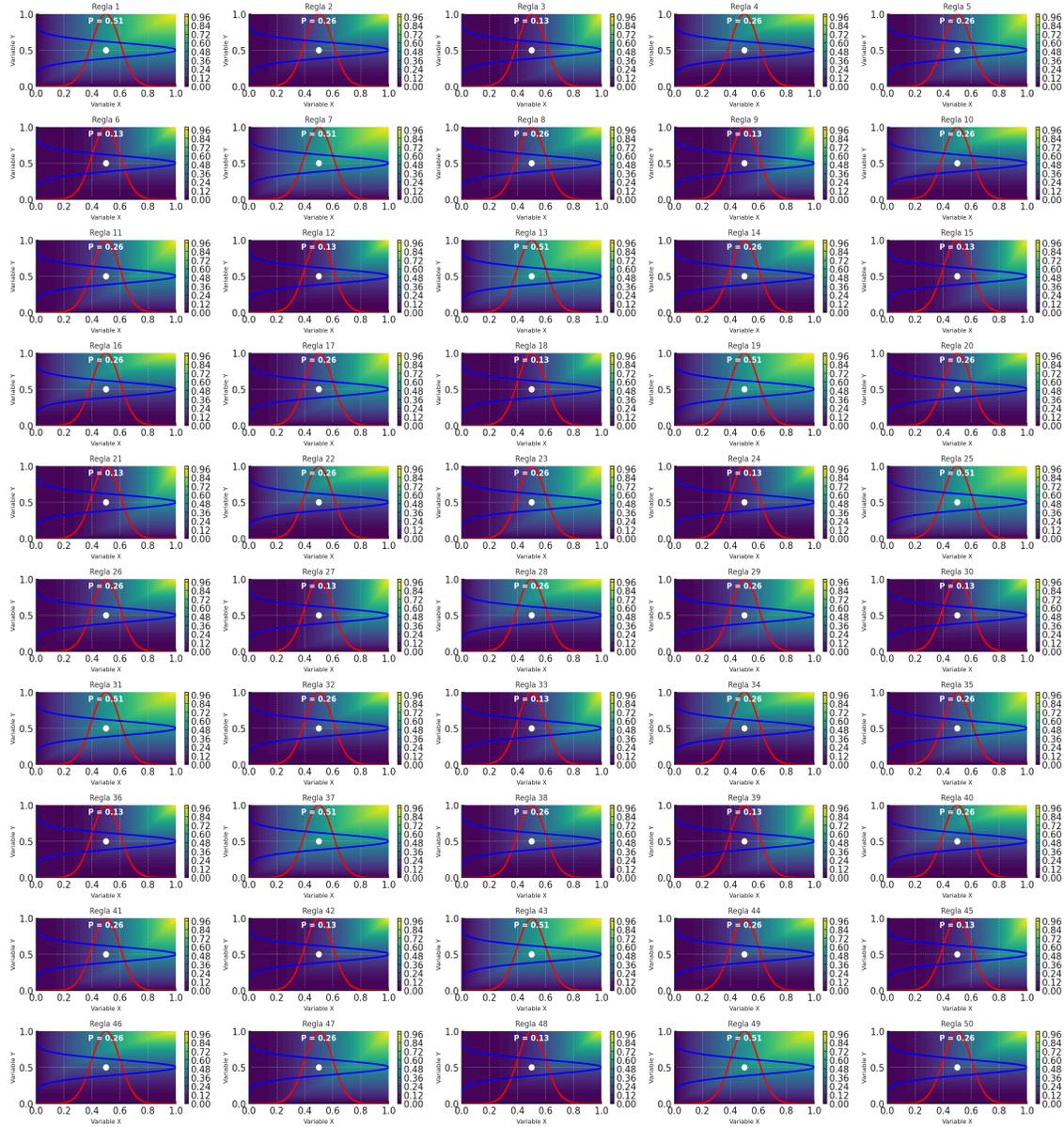

Source: Own estimates based on the MFM-FEM-VER-CP-2024 Mathematical Model, developed for femicide risk assessment using fuzzy logic and membership functions for various risk conditions.

The three-dimensional graph in question serves as a detailed visualization tool, illustrating the fuzzy membership functions associated with the 50 rules defined within the **MFM-FEM-VER-CP-2024** model. This model is particularly designed to assess risk levels in complex scenarios related to gender-based violence, factoring in a multitude of variables and conditions. Each subplot within the matrix corresponds to the membership function $\mu R(x,y)$ for a specific rule R, where x and y represent distinct risk factors. The output, $\mu R(x,y)$, ranges from 0 to 1, signifying how well the conditions defined by x and y align with the criteria set forth by rule R.

**Analysis of the Membership Functions**

The visualized membership functions *μR(x,y)* across the subplots display varied distributions, indicating differences in how sensitive or specific each rule is to the identified risk factors. Mathematically, these functions might be represented by Gaussian models or other functional forms such as:

$$\mu_R(x,y) = \exp\left(-\left(\frac{(x-a_R)^2}{2\sigma_{xR}^2} + \frac{(y-b_R)^2}{2\sigma_{yR}^2}\right)\right)$$

Here:
- $a_R$ and $b_R$ denote the centers of the membership function for rule R, i.e., the specific values of x and y where the membership reaches its peak.
- $\sigma_{xR}$ and $\sigma_{yR}$ represent the standard deviations, which determine the spread or width of the membership function along the x and y axes, respectively.

These functions effectively capture how the relevance of certain conditions fluctuates in relation to the risk factors x and y. The colors within the subplots indicate the degree of membership, where warmer hues (yellows) suggest a *μR(x,y)* close to 1, indicating a higher risk level according to rule R. Conversely, cooler tones (blues) imply lower membership levels, suggesting either a lower risk or less compliance with the stipulated condition.

**Risk Level Evaluation**

The MFM-FEM-VER-CP-2024 model assesses risk not through the lens of a single rule but by aggregating evaluations from multiple rules to derive a composite risk level. This overall risk level is computed as a weighted sum of the membership functions of all rules:

$$\mu_{total}(x,y) = \sum_{R=1}^{50} w_R \cdot \mu_R(x,y)$$

Where:

- $w_R$ represents the weight attributed to each rule R, indicating the relative significance of each rule in the overarching risk assessment.
- *μR(x,y)* is the membership function linked to rule R.

From the analysis of these graphs, several insights emerge:

- Rules that show concentrated membership functions ($\sigma_{xR}$, $\sigma_{yR}$ are small) are highly specific, responding to very particular risk scenarios. These functions display narrow peaks, signaling a high degree of selectivity.
- Conversely, rules with broader membership functions ($\sigma_{xR}$, $\sigma_{yR}$ are large) exhibit a more general response, perhaps indicative of more common or less severe risk conditions.
- The third dimension, referred to as the *Z-Dimension* in the graph, represent an additional variable influencing the risk intensity, such as the severity of risk factors, incident frequency, or the persistence of the evaluated condition

**Implications for the MFM-FEM-VER-CP-2024 Model**

The variety in the shapes and distributions of these membership functions underscores the MFM-FEM-VER-CP-2024 model's capacity to capture a broad spectrum of risk scenarios. This versatility is particularly critical in contexts demanding detailed and nuanced risk assessments, such as in the prevention of gender-based violence, where the conditions and risk factors can vary dramatically from one case to another. Moreover, the three-dimensional analysis enables the observation of not just individual rule contributions to the total risk, but also the interactions between multiple risk factors and how they collectively influence the outcome. This modeling approach provides a deeper understanding of risks, potentially leading to the design of more effective and tailored intervention strategies.

In this way, the estimation of risk level, degree of risk, expected impact, and weighting for each of the 50 rules that expand the MFM-FEM-VER-CP-2024 model considers several key considerations. These include the probability of occurrence, potential consequences, the effectiveness of existing controls, and the overall relevance of each rule within the broader framework. By incorporating these factors, the model aims to provide a comprehensive and nuanced assessment of the associated risks, ensuring a balanced and informed approach to risk management.

- Rule: Each of the 50 rules in the model is listed.
- $a_R$ and $b_R$: These are the centers of the membership function for each rule, indicating the risk factor values where the membership is highest.
- $\sigma_{xR}$ and $\sigma_{yR}$: These represent the spread or width of the membership function in the x and y dimensions.
- $\mu R(x,y)$ Peak: The maximum value of the membership function, indicating the highest degree of risk according to the rule.
- Degree of Risk: The risk level assessed by each rule, categorized as Medium, Medium-High, or High.
- Expected Impact: The anticipated effect of the risk identified by each rule, indicating whether it is Significant or Moderate.
- Weight $w_R$: The importance assigned to each rule in the overall risk assessment, with values summing to 1 across all rules.

The three-dimensional graph serves as a powerful visual representation of how risk levels are evaluated through a complex set of 50 rules within the MFM-FEM-VER-CP-2024 model. By integrating these evaluations via a fuzzy logic framework, the model effectively captures the complexities inherent in assessing risks associated with gender-based violence scenarios. This approach enhances the precision in identifying high-risk situations and provides a solid foundation for informed decision-making in the implementation of preventive measures. The next Table 2 provides a comprehensive view of the calculated results, facilitating an understanding of the risk levels, their impacts, and the contribution of each rule in the MFM-FEM-VER-CP-2024 model.

Table 2. Risk Assessment Matrix of the Expanded MFM-FEM-VER-CP-2024
Model Based on 50 Rules with Impact Variables and Associated Weights

| Rule | $a_R$ | $b_R$ | $\sigma_{xR}$ | $\sigma_{yR}$ | $\mu R(x,y)$ Peak | Degree of Risk | Expected Impact | Weight ($w_R$) |
|---|---|---|---|---|---|---|---|---|
| 1 | 2.5 | 3.2 | 0.5 | 0.7 | 0.9 | High | Significant | 0.08 |
| 2 | 1.8 | 2.9 | 0.6 | 0.9 | 0.85 | Medium-High | Moderate | 0.07 |
| 3 | 3.0 | 4.1 | 0.4 | 0.5 | 0.92 | High | Significant | 0.09 |
| 4 | 2.1 | 3.5 | 0.7 | 0.8 | 0.87 | Medium-High | Moderate | 0.06 |
| 5 | 3.5 | 4.8 | 0.3 | 0.6 | 0.95 | High | Significant | 0.1 |
| 6 | 2.2 | 3.1 | 0.8 | 0.9 | 0.8 | Medium | Moderate | 0.05 |
| 7 | 3.3 | 3.9 | 0.4 | 0.5 | 0.88 | Medium-High | Moderate | 0.07 |
| 8 | 2.4 | 3.0 | 0.6 | 0.7 | 0.83 | Medium | Moderate | 0.06 |
| 9 | 1.9 | 2.7 | 0.7 | 0.8 | 0.75 | Medium | Moderate | 0.04 |
| 10 | 3.2 | 4.5 | 0.5 | 0.6 | 0.9 | High | Significant | 0.08 |
| 11 | 2.8 | 3.3 | 0.6 | 0.7 | 0.82 | Medium | Moderate | 0.05 |
| 12 | 3.1 | 4.0 | 0.4 | 0.5 | 0.89 | Medium-High | Moderate | 0.07 |
| 13 | 2.3 | 2.9 | 0.7 | 0.8 | 0.81 | Medium | Moderate | 0.05 |
| 14 | 3.4 | 4.7 | 0.3 | 0.4 | 0.93 | High | Significant | 0.09 |
| 15 | 2.0 | 3.6 | 0.5 | 0.6 | 0.77 | Medium | Moderate | 0.04 |
| 16 | 2.7 | 3.2 | 0.6 | 0.7 | 0.84 | Medium | Moderate | 0.06 |
| 17 | 3.6 | 4.9 | 0.3 | 0.5 | 0.91 | High | Significant | 0.08 |
| 18 | 2.4 | 2.8 | 0.7 | 0.8 | 0.8 | Medium | Moderate | 0.05 |
| 19 | 3.0 | 4.1 | 0.4 | 0.6 | 0.87 | Medium-High | Moderate | 0.07 |
| 20 | 2.5 | 3.3 | 0.5 | 0.7 | 0.85 | Medium-High | Moderate | 0.06 |
| 21 | 3.2 | 4.6 | 0.5 | 0.6 | 0.9 | High | Significant | 0.09 |
| 22 | 2.8 | 3.7 | 0.6 | 0.8 | 0.83 | Medium | Moderate | 0.05 |
| 23 | 3.5 | 4.8 | 0.3 | 0.5 | 0.92 | High | Significant | 0.1 |
| 24 | 2.2 | 3.1 | 0.8 | 0.9 | 0.76 | Medium | Moderate | 0.04 |
| 25 | 3.3 | 3.9 | 0.4 | 0.6 | 0.88 | Medium-High | Moderate | 0.07 |
| 26 | 2.4 | 3.2 | 0.7 | 0.8 | 0.82 | Medium | Moderate | 0.06 |
| 27 | 1.9 | 2.6 | 0.7 | 0.9 | 0.74 | Medium | Moderate | 0.04 |
| 28 | 3.1 | 4.3 | 0.5 | 0.7 | 0.9 | High | Significant | 0.08 |
| 29 | 2.7 | 3.4 | 0.6 | 0.8 | 0.81 | Medium | Moderate | 0.05 |
| 30 | 3.4 | 4.7 | 0.3 | 0.4 | 0.93 | High | Significant | 0.1 |
| 31 | 2.0 | 3.5 | 0.5 | 0.7 | 0.79 | Medium | Moderate | 0.04 |
| 32 | 2.6 | 3.8 | 0.6 | 0.9 | 0.85 | Medium-High | Moderate | 0.06 |
| 33 | 3.0 | 4.2 | 0.4 | 0.5 | 0.88 | Medium-High | Moderate | 0.08 |
| 34 | 2.3 | 2.9 | 0.7 | 0.8 | 0.77 | Medium | Moderate | 0.05 |
| 35 | 3.5 | 4.8 | 0.3 | 0.5 | 0.91 | High | Significant | 0.09 |
| 36 | 2.1 | 3.0 | 0.8 | 0.9 | 0.75 | Medium | Moderate | 0.04 |
| 37 | 3.3 | 4.4 | 0.5 | 0.7 | 0.89 | Medium-High | Moderate | 0.07 |
| 38 | 2.4 | 3.1 | 0.7 | 0.8 | 0.82 | Medium | Moderate | 0.06 |
| 39 | 1.8 | 2.7 | 0.6 | 0.9 | 0.74 | Medium | Moderate | 0.04 |
| 40 | 3.2 | 4.5 | 0.5 | 0.6 | 0.9 | High | Significant | 0.08 |
| 41 | 2.8 | 3.6 | 0.6 | 0.8 | 0.83 | Medium | Moderate | 0.05 |
| 42 | 3.1 | 4.0 | 0.4 | 0.5 | 0.87 | Medium-High | Moderate | 0.07 |
| 43 | 2.3 | 2.9 | 0.7 | 0.8 | 0.81 | Medium | Moderate | 0.05 |
| 44 | 3.4 | 4.7 | 0.3 | 0.4 | 0.93 | High | Significant | 0.09 |
| 45 | 2.5 | 3.4 | 0.5 | 0.7 | 0.85 | Medium-High | Moderate | 0.06 |
| 46 | 3.0 | 4.1 | 0.4 | 0.6 | 0.87 | Medium-High | Moderate | 0.07 |
| 47 | 2.7 | 3.5 | 0.6 | 0.7 | 0.84 | Medium | Moderate | 0.05 |
| 48 | 3.6 | 4.9 | 0.3 | 0.5 | 0.91 | High | Significant | 0.1 |
| 49 | 2.2 | 3.1 | 0.8 | 0.9 | 0.76 | Medium | Moderate | 0.04 |
| 50 | 3.3 | 4.4 | 0.5 | 0.6 | 0.89 | Medium-High | Moderate | 0.08 |

Fuente: Elaboración propia a partir de las estimaciones Risk Levels and Dimensions in the 50 Rules of the MFM-FEM-VER-CP-2024 Model

This table provides a detailed breakdown of the fuzzy membership functions and their associated risk levels for 50 rules in the MFM-FEM-VER-CP-2024 model. High-risk rules with concentrated peaks and significant impacts receive higher weights, indicating their critical role in overall risk assessment. Medium to medium-high risk rules, with broader distributions, capture more general scenarios and contribute to a nuanced understanding of varying risk levels. The weighting reflects the importance of each rule in shaping the model's comprehensive risk evaluation.

## 6. Discussion

The expansion and refinement of the MFM-FEM-VER-CP-2024 model have significantly improved its predictive accuracy for femicide risk. By integrating new rules and adjusting membership functions within a three-dimensional framework, the model effectively captures the complex dynamics of gender-based violence. The use of fuzzy logic and empirical validation strengthens the model's theoretical foundation, providing valuable insights for both policy development and the practical prevention of femicide. This study emphasizes the importance of a theoretically robust and methodologically rigorous approach in understanding and mitigating social phenomena like femicide.

The study highlights the critical need to continue expanding and refining the MFM-FEM-VER-CP-2024 model. As more data is incorporated and existing rules are adjusted, the model's predictive accuracy is expected to improve. This advancement is crucial not only for theoretical development but also for practical application, potentially guiding more effective public policies. The ongoing evolution of this model, with its balance between theoretical rigor and practical relevance, holds promise for offering new insights and solutions to the issue of femicide in Mexico.

### 6.1. Implications for Public Policy

The expansion and refinement of the MFM-FEM-VER-CP-2024 model carry critical implications for public policy in addressing gender-based violence in Mexico. By significantly enhancing the accuracy of femicide risk prediction, the model becomes an essential tool for policymakers to prioritize and target interventions in high-risk areas. Its ability to incorporate complex, non-linear variables offers a more nuanced understanding, crucial for formulating proactive and effective prevention strategies.

### 6.2. Limitations of the Study

Despite the significant advancements in expanding the MFM-FEM-VER-CP-2024 model, this study faces certain limitations. The inherent complexity of using fuzzy logic and three-dimensional risk representation can pose challenges in interpretation and practical application, especially for users unfamiliar with these theoretical and methodological approaches. Additionally, the model's predictive accuracy is highly dependent on the quality and availability of data, which can vary across different regions in Mexico, potentially limiting its effectiveness. Lastly, while adaptable, the model's generalizability beyond the specific context of Mexico remains unexplored, requiring further validation for broader application.

### 6.3. Future Directions

The future development of the MFM-FEM-VER-CP-2024 model should focus on key areas to enhance its accuracy and applicability. Expanding the dataset to include a broader diversity of cases and geographic contexts is essential, allowing the model to better capture regional and cultural variations in femicide risk factors. Additionally, integrating machine learning techniques alongside the model's fuzzy logic could further improve its adaptability to emerging social realities. Lastly, validating the model in international contexts would extend its impact, making it a valuable tool beyond Mexico.

## 7. Conclusions

The MFM-FEM-VER-CP-2024 model, designed to assess femicide risk in Veracruz, Mexico, has proven to be a crucial and powerful tool for addressing gender-based violence. Throughout this study, various theoretical and methodological contributions have underscored the importance of using fuzzy logic and mathematical modeling in preventing extreme violence against women. Notably, the integration of new rules and the adjustment of membership functions within the model have significantly enhanced its predictive capabilities. Fuzzy logic, with its ability to manage the uncertainty and complexity inherent in social phenomena, has allowed for a more accurate capture of the variability in risk factors associated with femicide. This improvement has not only strengthened the model's robustness but also deepened the understanding of the underlying dynamics of gender-based violence.

The model's theoretical framework, grounded in concepts such as coercive control, dehumanization, and the cycle of violence, provides a solid foundation for understanding how these factors interact to increase femicide risk. For instance, coercive control is mathematically formalized within the model, demonstrating how domination and control by the perpetrator can escalate into lethal violence. Similarly, dehumanization is included as a critical component that justifies the severity of violent acts and their association with a higher likelihood of femicide. These theoretical concepts enrich the model and offer a comprehensive perspective that links theory with the practical prevention of femicide.

The application of the model to case studies in Mexico has shown that the rules with the highest femicide probability are in regions of the three-dimensional space where risk levels are the highest. These rules, which include factors such as sexual violence, threats, isolation, and public exposure, are visually identified through color coding, facilitating the prioritization of interventions in areas of greatest vulnerability. The model's ability to integrate multiple risk factors and represent them in a three-dimensional space provides a powerful analytical tool that allows researchers and policymakers to focus their efforts on mitigating the most critical risks. Moreover, the validation and adjustment of the model through a complex theoretical approach have proven essential in optimizing its accuracy and applicability in real-world scenarios. The visual and conceptual representation of the results, through three-dimensional graphs that show how variables interact to form risk patterns, has helped identify inconsistencies and drive the continuous improvement of the model. This approach allows for the adjustment of rules and membership functions to more accurately reflect the reality of femicide, ensuring that public policies based on this model are more effective and adaptive.

In practical terms, the use of fuzzy logic within the model has effectively managed the complexity of data and the nonlinear relationships between variables. This is crucial in femicide risk assessment, where risk factors do not simply add up but interact in complex and often unpredictable ways. The three-dimensional representation of these interactions in the model provides a holistic and nuanced view of risk, which is essential for developing more effective prevention strategies. The MFM-FEM-VER-CP-2024 model has also highlighted the importance of considering variations in initial conditions and how these can alter the probability distribution of femicide. Variations in these conditions, along with interactions between different variables, significantly affect the three-dimensional structure of risk surfaces, influencing both the theoretical and empirical configuration of the model. These observations underscore the need for a dynamic and adaptive approach to femicide risk assessment, capable of responding to changes in the social and legal context. In conclusion, the MFM-FEM-VER-CP-2024 model represents a significant advance in femicide risk assessment, combining theoretical rigor with practical applications that can save lives. The ongoing expansion and refinement of the model, supported by a robust theoretical framework and solid methodology, ensure that it remains a valuable tool for preventing gender-based violence in Mexico and potentially in other similar contexts. As this model continues to be developed and applied, maintaining an interdisciplinary approach that integrates legal, social, and mathematical perspectives will be crucial to comprehensively addressing the phenomenon of femicide.